\documentclass{article}
\usepackage{ijcai20}

\usepackage{times}
\usepackage{helvet}
\usepackage{courier}
\usepackage{url}
\usepackage{amsmath}
\usepackage{amssymb}
\usepackage{graphicx}
\usepackage{listings}
\usepackage[utf8]{inputenc}
\usepackage{graphicx}

\usepackage{color}
\usepackage{algorithm}

\def\mvis{\!=\!}

\def\bi{\begin{itemize}}
\def\ii{\item}
\def\ei{\end{itemize}}
\def\beq{\begin{equation}}
\def\eeq#1{\label{#1}\end{equation}}
\def\ba{\begin{array}}
\def\ea{\end{array}}
\def\i#1{\hbox{\it #1\/}}
\def\mi#1{\mathit{#1}}

\def\lpmln{\hbox{\rm LP}^{\rm{MLN}}}
\def\lpmln{{\rm LP}^{\rm{MLN}}}
\def\no{\i{not}}
\def\sneg{\sim\!\!}
\def\ar{\leftarrow}

\def\no{\i{not}}
\def\false{\hbox{\sc false}}
\def\true{\hbox{\sc true}}

\def\i#1{\hbox{\itshape #1\/}}

\def\nasp{{\rm NeurASP}}

\newcommand{\argmax}{\mathop{\mathrm{argmax}}\limits}

\newtheorem{prop}{Proposition}

\newtheorem{example}{Example}

\newcommand{\cblu}{\color{black}}
\newcommand{\cred}{\color{red}}

\long\def\BOC#1\EOC{\message{(Commented text )}}
\long\def\BOCC#1\EOCC{\message{(Commented text )}}
\long\def\BOCCC#1\EOCCC{\message{(Commented text )}}
\long\def\optional#1{\empty}
\long\def\NB#1{}
\long\def\NBB#1{}

\title{$\nasp$: Embracing Neural Networks into Answer Set Programming}

\author{
Zhun Yang$^1$ \and
Adam Ishay$^1$ \And
Joohyung Lee$^1$ $^2$\ \footnote{Contact Author} \\
\affiliations
$^1$ Arizona State University, Tempe, AZ, USA \\
$^2$ Samsung Research, Seoul, South Korea
\emails
\texttt{\{zyang90,aishay,joolee\}@asu.edu}
}

\begin{document}

\maketitle

\begin{abstract}
We present $\nasp$, a simple extension of answer set programs by embracing neural networks. By treating the neural network output as the probability distribution over atomic facts in answer set programs, $\nasp$ provides a simple and effective way to integrate sub-symbolic and symbolic computation. We demonstrate how $\nasp$ can make use of a pre-trained neural network in symbolic computation and how it can improve the neural network's perception result by applying symbolic reasoning in answer set programming.  Also, $\nasp$ can be used to train a neural network better by training with ASP rules so that a neural network not only learns from implicit correlations from the data but also from the explicit complex semantic constraints expressed by the rules. 
\end{abstract}

\section{Introduction} \label{sec:intro}

The integration of low-level perception with high-level reasoning is one of the oldest problems in Artificial Intelligence. Today, the topic is revisited with the recent rise of deep neural networks. Several proposals were made to implement the  reasoning process in complex neural network architectures, e.g., \cite{kathryn18tensorlog,rocktaschel17end,donadello17logic,kazemi18relnn,sourek15lifted,palm18recurrent,lin19kagnet}. 
However, it is still not clear how complex and high-level reasoning, such as default reasoning \cite{rei80}, ontology reasoning \cite{baader03handbook}, and causal reasoning \cite{pearl00causality}, can be successfully computed by these approaches. 
The latter subject has been well-studied in the area of knowledge representation (KR), but many KR formalisms, including answer set programming (ASP) \cite{lif08,bre11}, are logic-oriented and do not incorporate high-dimensional vector space and pre-trained models for perception tasks as handled in deep learning, which limits the applicability of KR in many practical applications involving data and uncertainty.

In this paper, we present a simple extension of answer set programs by embracing neural networks. Following the idea of DeepProbLog \cite{manhaeve18deepproblog}, by treating the neural network output as the probability distribution over atomic facts in answer set programs, the proposed $\nasp$ provides a simple and effective way to integrate  sub-symbolic and symbolic computation. 

We demonstrate how $\nasp$ can be useful for some tasks where both perception and reasoning are required. 
Reasoning can help identify perception mistakes that violate semantic constraints, which in turn can make perception more robust. 
For example, a neural network for object detection may return a bounding box and its classification ``car,'' but it may not be clear whether it is a real car or a toy car. The distinction can be made by applying reasoning about the relations with the surrounding objects and using commonsense knowledge. 
Or when it is unclear whether a round object attached to the car is a wheel or a doughnut, the reasoner could conclude that it is more likely to be a wheel by applying commonsense knowledge. 
In the case of a neural network that recognizes digits in a given Sudoku board, the neural network may get confused if a digit next to $1$ in the same row is $1$ or $2$, but the reasoner can conclude that it cannot be $1$ by applying the constraints for Sudoku. 

Another benefit of this hybrid approach is that it alleviates the burden of neural networks when the constraints/knowledge are already given.  
Instead of building a large end-to-end neural network that learns to solve a Sudoku puzzle given as an image, we can let a neural network only do digit recognition and use ASP to find the solution of the recognized board. This makes the design of the neural network simpler and the required training dataset much smaller. Also, when we need to solve some variation of Sudoku, such as Anti-knight or Offset Sudoku, the modification is simpler than training another large neural network from scratch to solve the new puzzle.


$\nasp$ can also be used to train a neural network together with rules so that a neural network not only learns from implicit correlations from the data but also from explicit complex semantic constraints expressed by ASP rules. 
The {\em semantic loss} \cite{xu18asemantic} obtained from the reasoning module can be backpropagated into the rule layer and then further into neural networks via neural atoms. 
This sometimes makes a neural network learn better even with fewer data. 

Compared to DeepProbLog, $\nasp$ supports a rich set of KR constructs supported by answer set programming that allows for convenient representation of complex knowledge. It utilizes an ASP solver in computation instead of constructing circuits as in DeepProbLog. 

The paper is organized as follows. Section~\ref{sec:nasp} introduces the syntax and the semantics of $\nasp$. Section~\ref{sec:inference} illustrates how reasoning in $\nasp$ can enhance the perception result by considering relations among objects perceived by pre-trained neural networks. Section~\ref{sec:learning} presents learning in $\nasp$ where ASP rules work as a semantic regularizer for training neural networks so that neural networks are trained not only from data but also from rules. 
Section~\ref{sec:related} examines related works and Section~\ref{sec:conclusion} concludes.

The implementation of $\nasp$, as well as codes used for the experiments, is publicly available online at 
\begin{center}
\url{https://github.com/azreasoners/NeurASP}.\par
\end{center}

\section{$\nasp$} \label{sec:nasp}

We present the syntax and the semantics of $\nasp$.

\subsection{Syntax} \label{ssec:syntax}

We assume that neural network $M$ allows an arbitrary tensor as input whereas the output is a matrix in $\mathbb{R}^{e\times n}$, 
where $e$ is the number of random events predicted by the neural network and $n$ is the number of possible outcomes for each random event. Each row of the matrix represents the probability distribution of the outcomes of each event. 
For example, if $M$ is a neural network for MNIST digit classification, then the input is a tensor representation of a digit image, $e$ is $1$, and $n$ is $10$. If $M$ is a neural network that outputs a Boolean value for each edge in a graph, then $e$ is the number of edges and $n$ is $2$. 
Given an input tensor ${\bf t}$, by $M({\bf t})$, we denote the output matrix of $M$.
The value $M({\bf t})[i,j]$ (where $i\in\{1,\dots, e\}$, $j\in\{1,\dots,n\}$) is the probability of the $j$-th outcome of the $i$-th event upon the input ${\bf t}$.

In $\nasp$, the neural network $M$ above can be represented by a {\em neural atom} of the form 
\beq
   nn(m(e, t), \left[v_1, \dots, v_n \right]), 
\eeq{eq:nn-atom}
where
(i) $nn$ is a reserved keyword to denote a neural atom; 
(ii) $m$ is an identifier (symbolic name) of the neural network $M$;
(iii) $t$ is a list of terms that serves as a ``pointer'' to an input data;  related to it, there is a mapping ${\bf D}$ (implemented by an external Python code) that turns $t$ into an input tensor; 
%
(iv) $v_1, \dots, v_n$ represent all $n$ possible outcomes of each of the $e$ random events.

Each neural atom \eqref{eq:nn-atom} introduces propositional atoms of the form $c\mvis v$, where $c\in\{m_1(t), \dots, m_e(t)\}$ and $v\in\{v_1,\dots,v_n\}$. The output of the neural network provides the probabilities of the introduced atoms (defined in Section~\ref{ssec:semantics}).

\begin{example} \label{ex:digit}
Let $M_{digit}$ be a neural network that classifies an MNIST digit image. The input of $M_{digit}$ is (a tensor representation of) an image and the output is a matrix in $\mathbb{R}^{1\times 10}$. 
The neural network can be represented by the neural atom
\[
   nn( digit(1,d),\  [0,1,2,3,4,5,6,7,8,9]),
\]
which introduces propositional atoms $digit_1(d)\mvis 0$, $digit_1(d)\mvis 1$, $\dots$, $digit_1(d)\mvis 9$.
\end{example}

\begin{example}
Let $M_{sp}$ be another neural network for finding the shortest path in a graph with 24 edges. The input is a tensor encoding the graph and the start/end nodes of the path, and the output is a matrix in $\mathbb{R}^{24\times 2}$.
This neural network can be represented by the neural atom
\[
    nn( sp(24,g),\  [\true, \false]) .
\]
\end{example}

A {\em $\nasp$ program} $\Pi$ is the union of $\Pi^{asp}$ and $\Pi^{nn}$, where $\Pi^{asp}$ is a set of propositional rules (standard rules as in ASP-Core 2 \cite{calimeri20asp}) and $\Pi^{nn}$ is a set of neural atoms.  Let $\sigma^{nn}$ be the set of all atoms $m_i(t) \mvis v_j$ that is obtained from the neural atoms in $\Pi^{nn}$  as described above. We require that, in each rule $\i{Head}\ar\i{Body}$ in 
$\Pi^{asp}$, no atoms in $\sigma^{nn}$ appear in $\i{Head}$.

We could allow schematic variables into $\Pi$, which are understood in terms of grounding as in standard answer set programs.  We find it convenient to use rules of the form 
\beq
   nn(m(e, t), \left[v_1, \dots, v_n \right]) \ar \i{Body}
\eeq{eq:nn-rule}
where $\i{Body}$ is either identified by $\top$ or $\bot$ during grounding so that \eqref{eq:nn-rule} can be viewed as an abbreviation of multiple (variable-free) neural atoms~\eqref{eq:nn-atom}.

\begin{example}\label{ex:addition}
An example $\nasp$ program $\Pi_{digit}$ is as follows, where  $d_1$ and $d_2$ are terms representing two images. Each image is classified by neural network $M_{digit}$ as one of the values in $\{0,\dots,9 \}$. The addition of two digit-images is the sum of their values. 
\beq
{
\ba {l}
img(d_1). \\
img(d_2).\\[0.3em]
nn(digit(1, X), [0,1,2,3,4,5,6,7,8,9]) \leftarrow img(X).\\[0.3em]
addition(A,B,N) \leftarrow
    digit_1(A) \mvis N_1, 
    digit_1(B) \mvis N_2, \\
\hspace{3.3cm} N=N_1+N_2.
\ea
}
\eeq{digit}
The neural network $M_{digit}$ outputs 10 probabilities for each image. The addition is applied once the digits are recognized and its probability is induced from the perception as we explain in the next section. 
\end{example}

\subsection{Semantics} \label{ssec:semantics}
{\cblu 
For any $\nasp$ program $\Pi=\Pi^{asp}\cup \Pi^{nn}$, we first obtain its ASP counterpart 
$\Pi'=\Pi^{asp}\cup \Pi^{ch}$
where $\Pi^{ch}$ consists of the following set of rules for each neural atom \eqref{eq:nn-atom} in $\Pi^{nn}$
\[ 
\ba {l}
    \{m_i(t) \mvis v_1;~ \dots ~; m_i(t) \mvis v_n\} = 1   \ \ \ \ \text{for } i\in \{1,\dots e\}.
\ea
\]
}
The above rule (in the language of {\sc clingo}) means to choose exactly one atom in between the set braces.\footnote{In practice, each atom $m_i(t)=v$ is written as $m(i,t,v)$.}
We define the {\em stable models} of $\Pi$ as the stable models of $\Pi'$,
{\cblu 
and define the {\em total choices} of $\Pi$ as the stable models of $\Pi^{ch}$. For each total choice $C$ of $\Pi$, we use $\i{Num}(C,\Pi)$ to denote the number of stable models of $\Pi$ that satisfy $C$. We require a $\nasp$ program $\Pi$ to be {\em coherent} such that $\i{Num}(C,\Pi) > 0$ for every total choice $C$ of $\Pi$.
}

To define the probability of a stable model, we first define the probability of an atom $m_i(t)\mvis v_j$ in $\sigma^{nn}$. Recall that there is an external mapping ${\bf D}$ that turns $t$ into a specific input tensor of $M$.
The probability of each atom $m_i(t) \mvis v_j$ is defined as $M({\bf D}(t))[i,j]$: 
\[
   P_{\Pi}(m_i(t) \mvis v_j) = M({\bf D}(t))[i,j] .\;
\]

For instance, 
recall that the output matrix of $M_{digit}({\bf D}(d))$ in Example~\ref{ex:addition} is in $\mathbb{R}^{1\times 10}$.  
The probability of atom $digit_1(d)=k$ is 
$M_{digit}({\bf D}(d))[1,k\!+\!1]$.

Given an interpretation $I$, by $I|_{\sigma^{nn}}$, we denote the projection of $I$ onto $\sigma^{nn}$.  
{\cblu Since $I|_{\sigma^{nn}}$ is a total choice of $\Pi$, $\i{Num}(I|_{\sigma^{nn}},\Pi)$ is the number of stable models of $\Pi$ that agree with $I|_{\sigma^{nn}}$ on $\sigma^{nn}$.}

\medskip
The probability of a stable model $I$ of $\Pi$ is defined as the product of the probability of each atom $c=v$ in $I|_{\sigma^{nn}}$, divided by the number of stable models  of $\Pi$ that agree with $I|_{\sigma^{nn}}$ on $\sigma^{nn}$. 
That is, for any interpretation $I$, 
\[
P_{\Pi}(I) = \begin{cases}
\frac{\prod\limits_{c=v \in I|_{\sigma^{nn}}} P_{\Pi}(c=v)}{\i{Num}(I|_{\sigma^{nn}}, \Pi)}
 & \text{if $I$ is a stable model of $\Pi$;}\\
0 & \text{otherwise.}
\end{cases}
\]

An {\em observation} is a set of ASP constraints (i.e., rules of the form $~\bot \leftarrow \i{Body}$).
The probability of an observation $O$ is defined as
\[
  P_{\Pi}(O) = \sum\limits_{I\models  O}P_{\Pi}(I)
\]
($I\models O$ denotes that $I$ satisfies $O$).

The probability of the set $ {\bf O}=\{O_1,\dots, O_o \}$ of observations is defined as the product of the probability of each $O_i$:
\[
{
P_{\Pi}({\bf O}) = \prod_{O_i\in {\bf O}}P_{\Pi}(O_i).
}
\]

\medskip
\noindent{\bf Example~\ref{ex:addition} Continued }
The ASP program $\Pi_{digit}'$, which is the ASP counterpart of $\Pi_{digit}$, is obtained from \eqref{digit} by replacing the third rule with 
\[
\ba {rl}
& \{digit_1(d_1)\mvis 0;~ \dots ~;  digit_1(d_1)\mvis 9\}=1. \\
& \{digit_1(d_2)\mvis 0 ;~ \dots ~; digit_1(d_2)\mvis 9\}=1.
\ea
\]
\BOC
The ASP program $\Pi_{digit}'$, which is the ASP counterpart of $\Pi_{digit}$, is as follows. 
\[
\small{
\ba {rl}
\Pi_{digit}^{asp}: & img(d_1). \hspace{2cm} img(d_2). \\
&addition(A,B,N) \leftarrow  digit_1(A)\mvis N_1, digit_1(B)\mvis N_2,\\
&\hspace{3.05cm}N\mvis N_1+N_2. \\
\Pi_{digit}^{ch}: & \{digit_1(d_1)\mvis 0;~ \dots ~;  digit_1(d_1)\mvis 9\}=1. \\
& \{digit_1(d_2)\mvis 0 ;~ \dots ~; digit_1(d_2)\mvis 9\}=1.
\ea
}
\]
\EOC
The following are the stable models of $\Pi_{digit}$, i.e., the stable models of $\Pi_{digit}'$.
{\small 
\[
\ba {l}
I_1 = \{digit_1(d_1)\mvis 0, digit_1(d_2)\mvis 0, addition(d_1, d_2, 0), \dots \},  \\
I_2 = \{digit_1(d_1)\mvis 0, digit_1(d_2)\mvis 1, addition(d_1, d_2, 1), \dots \}, \\
I_3 = \{digit_1(d_1)\mvis 1, digit_1(d_2)\mvis 0, addition(d_1, d_2, 1), \dots \},  \\
\dots,  \\
I_{100} = \{digit_1(d_1)\mvis 9, digit_1(d_2)\mvis 9, addition(d_1, d_2, 18), \dots \}.
\ea
\]
}
\NB{what is in ....} 
Their probabilities are as follows:
{\small 
\[
\ba {l}
P_{\Pi}(I_1) = M_{digit}({\bf D}(d_1))[1,1] \times M_{digit}({\bf D}(d_2))[1,1], \\
\dots, \\
P_{\Pi}(I_{100}) = M_{digit}({\bf D}(d_1))[1,10] \times M_{digit}({\bf D}(d_2))[1,10].
\ea
\]
}

\noindent The probability of $O=\{\leftarrow \i{not}\ addition(d_1, d_2, 1) \}$ is 
\[
P_{\Pi}(O) = P_{\Pi}(I_2) + P_{\Pi}(I_3).
\]

\section{Inference with $\nasp$} \label{sec:inference}

We implemented $\nasp$ by integrating PyTorch \cite{adam17automatic} and {\sc clingo} \cite{gebser11potassco}.  
PyTorch takes care of neural network processing including data loading and mapping ${\bf D}$ that maps pointer terms in neural atoms to input tensors. Computing the probability of a stable model is done by calling {\sc clingo} and post-processing in Python.
This section illustrates how this integration can be useful in reasoning about relations among objects recognized by neural networks.

\subsection{Commonsense Reasoning about Image} 

Suppose we have a neural network $M_\mi{label}$ that outputs classes of objects in the bounding boxes that are already detected. 
The following rule asserts that the neural network $M_\mi{label}$ classifies the bounding box $B$ into one of $\{car, cat, person, truck, other\}$, where $B$ is at location $(X_1,Y_1,X_2,Y_2)$ in image $I$:
\[
\ba {l}
nn(label(1,I,B), [car, cat, person, truck, other]) \leftarrow \\
\hspace{4cm} box(I,B,X_1,Y_1,X_2,Y_2).
\ea
\]

\begin{figure}[h!]
\vspace{-4mm}
\caption{Reasoning about relations among perceived objects}
\begin{center}
\includegraphics[width=8cm]{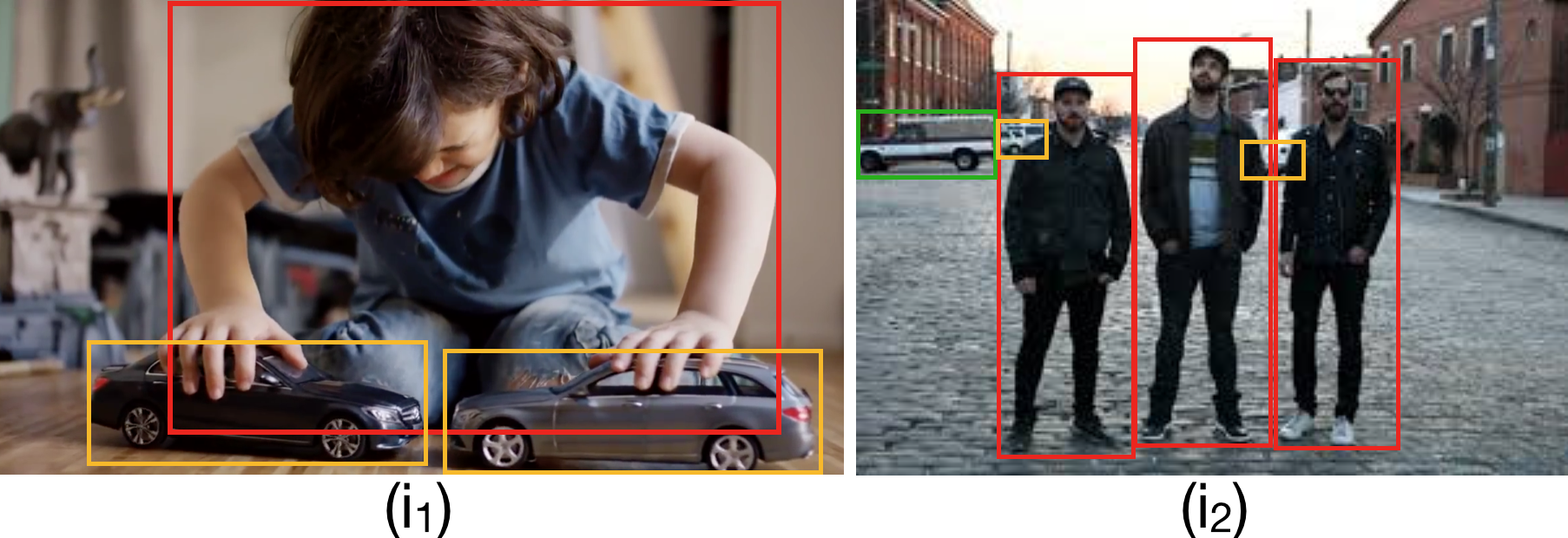}
\label{fig:yolo}
\end{center}
\vspace{-4mm}
\end{figure}

Consider the two images $i_1$ and $i_2$ in Figure~\ref{fig:yolo}.
The bounding boxes can be represented by the following facts. 
\[
\ba {l}
box(i_1, b_1, 100, 0, 450, 350).\\
box(i_1, b_2, 300, 300, 500, 400).\\
...
\ea
\]
The first rule says that there is a bounding box $b_1$ (i.e., the red box with a child) in image $i_1$, and the coordinates of its left-top and right-bottom corners are $(100,0)$ and $(450, 350)$.

Below we describe rules that allow for reasoning about the recognized objects. 
The following rules describe the general size relation between objects.
\[
\ba {l}
smaller(cat, person).\\ 
smaller(person, car).\\ smaller(person, truck).\\
smaller(X,Y) \leftarrow smaller(X,Z), smaller(Z,Y). 
\ea
\]

Next is the rule asserting that by default we conclude the same size relationship as above.
%
\[
{
\ba {l}
smaller(I, B_1, B_2) \leftarrow \i{not}\ \sneg smaller(I, B_1, B_2), \\
\hspace{0.5cm}label_1(I,B_1)\mvis L_1, label_1(I,B_2)\mvis L_2,
smaller(L_1,L_2).
\ea
}
\]
(The $\sneg$~ symbol stands for strong negation in ASP, which asserts explicit falsity.)

On the other hand, there are some exceptions, for instance, 
\[
{
\ba {l}
\sneg smaller(I, B_2, B_1) \leftarrow
         box(I,B_1,X_1,Y_1,X_2,Y_2),\\
\hspace{1.2cm} box(I,B_2,X_1',Y_1',X_2',Y_2'), ~Y_2 \geq Y_2', \\
\hspace{1.2cm} |X_1-X_2|\times |Y_1-Y_2| < |X_1'-X_2'|\times |Y_1'-Y_2'|.\\[1em]
smaller(I, B_1, B_2) \leftarrow\  \sneg smaller(I, B_2, B_1). \\[1em]

toy(I, B_1) \leftarrow
label_1(I,B_1)=L_1, ~label_1(I,B_2)=L_2,\\ 
\hspace{1.9cm} smaller(I, B_1, B_2), ~smaller(L_2,L_1). 
\ea
}
\]
The first rule says that ``$B_2$ is not smaller than $B_1$ if (i) $B_1$ and $B_2$ are objects in image $I$, (ii) $B_1$ is closer to the camera (i.e., $B_1$'s bottom boundary is closer to the bottom of $I$), and (iii) the box in the image for $B_1$ is smaller than $B_2$.'' \footnote{We assume that the camera is at the same height as the objects.}

The neural network model $M_\mi{label}$ outputs that the red boxes are persons, the yellow boxes are cars, and the green box is a truck. Upon this input and the rules above,  $\nasp$ allows us to derive that the two cars in image $i_1$ are {\em toy} cars, whereas the two cars in image $i_2$ are not: although they are surrounded by smaller boxes than those of humans, their boxes are not closer to the camera.

\subsection{Example: Solving Sudoku Puzzle in Image}  \label{subsec:sudoku_inference}

Consider the task of solving a Sudoku puzzle given as an image. In $\nasp$, we could use a neural network to recognize the digits in the given puzzle and use an ASP solver to compute the solution instead of having a single network that accounts for both perception and solving.

\BOCC
\begin{figure}[h!]
\caption{Reasoning on Sudoku}
\begin{center}
\includegraphics[width=8cm]{sudoku_pipeline.png}
\label{fig:sudoku_recognizer}
\end{center}
\vspace{-4mm}
\end{figure}
\BOC
Consider the Sudoku board in Figure~\ref{fig:sudoku_recognizer} (a).
\EOC
\EOCC

 We use the following $\nasp$ program $\Pi_\mi{sudoku}$ to first identify the digits in each grid cell on the board and then find the solution by assigning digits to all empty grid cells. 
\footnote{
The expression $~~\{a(R,C,N): N=1..9\}=1~~$ is a shorthand for $\{a(R,C,1); \dots;  a(R,C,9)\}=1~~$ in the language of {\sc clingo}.
%
}
\begin{lstlisting}
% identify the number in each of the 81 positions 
nn(identify(81, img), [empty,1,2,3,4,5,6,7,8,9]).

% assign one number N to each position (R,C)
a(R,C,N) :- identify(Pos,img,N), R=Pos/9, C=Pos\9, 
            N!=empty.
{a(R,C,N): N=1..9}=1 :- identify(Pos, img, empty), 
                        R=Pos/9, C=Pos\9.

% no number repeats in the same row
:- a(R,C1,N), a(R,C2,N), C1!=C2.

% no number repeats in the same column
:- a(R1,C,N), a(R2,C,N), R1!=R2.

% no number repeats in the same 3*3 box
:- a(R,C,N), a(R1,C1,N), R!=R1, C!=C1,
   ((R/3)*3 + C/3) = ((R1/3)*3 + C1/3).
\end{lstlisting}

The neural network model $M_\mi{identify}$ is rather simple. It is composed of 5 convolutional layers with dropout, a max pooling layer, and a $1\times 1$ convolutional layer followed by softmax. 
Given a Sudoku board image (.png file), neural network $M_\mi{identify}$ outputs a matrix in $\mathbb{R}^{81\times 10}$, which represents the probabilities of the values (empty, 1, \dots, 9) in each of the $81$ grid cells. 
The network $M_\mi{identify}$ is pre-trained using $\langle image, label\rangle$ pairs, where each $image$ is a Sudoku board image generated by {\em OpenSky Sudoku Generator} (\url{http://www.opensky.ca/~jdhildeb/software/sudokugen/}) and each $label$ is a vector of length 81 in which 0 is used to represent an empty cell at that position. 

Let $Acc_\mi{identify}$ denote the accuracy of identifying all empty cells and the digits on the board given as an image without making a single mistake in a grid cell. Let $Acc_{sol}$ denote the accuracy of solving a given Sudoku board without making a single mistake in a grid cell.
Let $r$ be the following rule in $\Pi_\mi{sudoku}$:
\begin{lstlisting}
{a(R,C,N): N=1..9}=1 :- identify(Pos, img, empty), 
                        R=Pos/9, C=Pos\9.
\end{lstlisting}
Table~\ref{tb:sudoku} compares $Acc_\mi{identify}$ of each of $M_\mi{identify}$,  $\nasp$ program $\Pi_\mi{sudoku}\!\setminus\!r$ with $M_\mi{identify}$, $\nasp$ program $\Pi_\mi{sudoku}$ with $M_\mi{identify}$, as well as 
$Acc_\mi{sol}$ of $\Pi_\mi{sudoku}$ with $M_\mi{identify}$.
\begin{table}[ht]
\caption{Sudoku: Accuracy on Test Data}
\centering
{\tiny
\begin{tabular} {c| c| c| c | c}
\hline
\hline
Num of  & $Acc_\mi{identify}$ of  & $Acc_\mi{identify}$ of  & $Acc_\mi{identify}$ of & $Acc_{sol}$ of  \\
Train Data  &  $M_\mi{identify}$   & $\nasp$ w/  & $\nasp$ w/ & $\nasp$ w/ \\
 & & $\Pi_\mi{sudoku}\!\setminus\! r$  & $\Pi_\mi{sudoku}$ & $\Pi_\mi{sudoku}$  \\
\hline
15  & 15\% & 49\% & 71\% & 71\% \\
17 & 31\% &  62\% & 80\% & 80\%  \\
19 & 72\% &  90\% & 95\% & 95\% \\
21 & 85\% & 95\% & 98\% & 98\% \\
23 & 93\% & 99\% & 100\% & 100\% \\
25 & 100\% & 100\% & 100\% & 100\% \\
\hline
\hline
\end{tabular}
\label{tb:sudoku}
}
\end{table}

Intuitively, $\Pi_\mi{sudoku}\!\setminus\! r$ only checks whether the identified numbers (by neural network $M_\mi{identify}$) satisfy the three constraints (the last three rules of $\Pi_\mi{sudoku}$), while $\Pi_\mi{sudoku}$ further checks whether there exists a solution given the identified numbers.
As shown in Table~\ref{tb:sudoku}, the use of reasoning in $\nasp$ program $\Pi_\mi{sudoku}\!\setminus\! r$ improves the accuracy $Acc_\mi{identify}$ of the neural network $M_\mi{identify}$ as explained in the introduction. 
The accuracy $Acc_\mi{identify}$ is further improved by trying to solve Sudoku completely using $\Pi_\mi{sudoku}$. 
\BOC
One example board that is not correctly identified by $M_\mi{identify}$ is in Figure~\ref{fig:sudoku_recognizer} (a), whose top 2 predictions by $M_\mi{identify}$, i.e., the $9\times 9$ values with the top 2 highest joint probabilities, are shown in Figure~\ref{fig:sudoku_recognizer} (b). Among them, the optimal prediction of $M_\mi{identify}$ contains two ``3'' at the last row thus does not satisfy the ASP program $\Pi_\mi{sudoku}'$. On the other hand, the second optimal prediction of $M_\mi{identify}$ satisfies $\Pi_\mi{sudoku}'$, thus is the final prediction of the $\nasp$ program $\Pi_\mi{sudoku}$ with the highest probability.
\EOC
Note that the solution accuracy $Acc_{sol}$ of $\Pi_\mi{sudoku}$ is equal  to the perception accuracy $Acc_\mi{identify}$ of $\Pi_\mi{sudoku}$ since the ASP yields a 100\% correct solution once the board is correctly identified.

\citeauthor{palm18recurrent} [\citeyear{palm18recurrent}] use a Graph Neural Network to solve Sudoku but the work restricts attention to textual input of the Sudoku board, not images as we do. Their work achieves 96.6\% accuracy after training with 216,000 examples. In comparison, even with the more challenging task of accepting images as input, the number of training examples we used is 15 -- 25, which is much less than the number of training examples used in~\cite{palm18recurrent}. Our work takes advantage of the fact that in a problem like Sudoku, where the constraints are explicitly given, a neural network only needs to focus on perception tasks, which is simpler than learning the perception and reasoning together. 

Furthermore, using the same trained perception neural network $M_\mi{identify}$, we can solve some elaborations of Sudoku problems by adding the following rules:

\medskip
\noindent{\bf [Anti-knight Sudoku]} No number repeats at a knight move

\begin{lstlisting}
:- a(R1,C1,N), a(R2,C2,N), |R1-R2|+|C1-C2|=3.
\end{lstlisting}

\noindent{\bf [Sudoku-X]} No number repeats at the  diagonals

\begin{lstlisting}
:- a(R1,C1,N), a(R2,C2,N), R1=C1, R2=C2, R1!=R2.
:- a(R1,C1,N), a(R2,C2,N), R1+C1=8, R2+C2=8, R1!=R2.
\end{lstlisting}

\BOC
\noindent{\bf [Offset Sudoku]} No number repeats at the same relative position in 3*3 boxes,
marked by same colors.
\begin{lstlisting}
:- a(R1,C1,N), a(R2,C2,N), R1\3 = R2\3, C1\3 = C2\3, 
   R1 != R2, C1 != C2.
\end{lstlisting}

\noindent{\bf [Hypersudoku]} No number repeats in 4 additional 3x3 boxes
\begin{lstlisting}
box(a, 1..3, 1..3).  box(b, 1..3, 5..7).
box(c, 5..7, 1..3).  box(d, 5..7, 5..7).
:- a(R1,C1,N), a(R2,C2,N), R1!=R2, C1!=C2,
   box(X,R1,C1), box(X,R2,C2).
\end{lstlisting}
\EOC

With neural network only approach, since the neural network needs to learn both perception and reasoning, each of the above variations would require training a complex and different model with a big dataset. 
However, with $\nasp$, the neural network only needs to recognize digits on the board. 
Thus solving each Sudoku variation above uses the same pre-trained model for the image input and we only need to add the aforementioned rules to $\Pi_\mi{sudoku}$.

Some Sudoku variations, such as Offset Sudoku, are in colored images. In this case, we need to increase the number of channels of $M_\mi{identify}$ from 1 to 3, and need to retrain the neural network with the colored images. Although not completely elaboration tolerant, compared to the pure neural network approach, this is significantly simpler. For instance, the number of training data needed to get 100\% perception accuracy for Offset Sudoku ($Acc_\mi{identify}$) is 70, which is still much smaller than what the end-to-end Sudoku solver would require. Using the new network trained, we only need to add the following rule to $\Pi_\mi{sudoku}$.

\medskip
\noindent{\bf [Offset Sudoku]} No number repeats at the same relative position in 3*3 boxes
\begin{lstlisting}
:- a(R1,C1,N), a(R2,C2,N), R1\3 = R2\3, 
   C1\3 = C2\3, R1 != R2, C1 != C2.
\end{lstlisting}

\BOCC
The results is shown in Table~\ref{tb:offsetsudoku} where $\Pi_{offset}$ is the union of $\Pi_\mi{sudoku}$ and the rule above for offset Sudoku.

\begin{table}[ht]
\caption{Colored Offset Sudoku: Accuracy on Test Data}
\centering
{\small 
\begin{tabular} {c| c| c| c | c}
\hline
\hline
Num of  & $Acc_\mi{identify}$ of  & $Acc_\mi{identify}$ of & $Acc_\mi{identify}$ of & $Acc_{sol}$ of  \\
Train Data & NN & $\Pi_{offset}\!\setminus\! r$  & $\Pi_{offset}$ & $\Pi_{offset}$  \\
\hline
15  & 5\% & 21\% & 51\% & 51\% \\
20 & 30\% &  53\% & 74\% & 74\%  \\
25 & 58\% &  71\% & 85\% & 85\% \\
30 & 71\% &  78\% & 89\% & 89\% \\
40 & 77\% & 86\% & 92\% & 92\% \\
50 & 94\% & 95\% & 97\% & 97\% \\
60 & 93\% & 95\% & 97\% & 97\% \\
70 & 100\% & 100\% & 100\% & 100\% \\
\hline
\hline
\end{tabular}
\label{tb:offsetsudoku}
}
\end{table}
\EOCC

\section{Learning in $\nasp$} \label{sec:learning}

We show how the semantic constraints expressed in $\nasp$ can be used to train neural networks better. 

\subsection{Gradient Ascent with $\nasp$} \label{ssec:learning}

In this section, we denote a $\nasp$ program by $\Pi({\boldsymbol \theta})$ where ${\boldsymbol \theta}$ is the set of the parameters in the neural network models associated with $\Pi$. Assume a $\nasp$ program $\Pi({\boldsymbol \theta})$ and a set ${\bf O}$ of observations such that $P_{\Pi({\boldsymbol \theta})}(O)>0$ for each $O \in {\bf O}$. 
The task is to find $\hat{\boldsymbol \theta}$ that maximizes the log-likelihood of observations ${\bf O}$ under program $\Pi({\boldsymbol \theta})$, i.e.,  
\[
\hat{\boldsymbol \theta} \in \argmax_{{\boldsymbol \theta}}~ log( P_{\Pi({\boldsymbol \theta})}({\bf O})), 
\]
which is equivalent to 
\[
\hat{\boldsymbol \theta} \in \argmax_{{\boldsymbol \theta}}~ \sum\limits_{O \in {\bf O}} log( P_{\Pi({\boldsymbol \theta})}(O)).
\]
Let ${\bf p}$ denote the probabilities of the atoms in $\sigma^{nn}$. Since ${\bf p}$ is indeed the outputs of the neural networks in $\Pi({\boldsymbol \theta})$, we can compute the gradient of ${\bf p}$ w.r.t. ${\boldsymbol \theta}$ through backpropagation. 
Then the gradient of $\sum\limits_{O \in {\bf O}} log( P_{\Pi({\boldsymbol \theta})}(O))$ w.r.t. ${\boldsymbol \theta}$ is
\[
\ba {rl}
\frac{\partial \sum\limits_{O \in {\bf O}} log( P_{\Pi({\boldsymbol \theta})}(O))}{\partial {\boldsymbol \theta}} 
& = \sum\limits_{O \in {\bf O}} \frac{\partial log(P_{\Pi({\boldsymbol \theta})}(O))}{\partial {\bf p}} \times \frac{\partial {\bf p}}{\partial {\boldsymbol \theta}}
\ea
\]
where  $\frac{\partial {\bf p}}{\partial {\boldsymbol \theta}}$ can be computed through the usual neural network backpropagation, while $\frac{\partial log(P_{\Pi({\boldsymbol \theta})}(O))}{\partial p}$ for each $p \in {\bf p}$ can be computed as follows. 

\begin{prop}\label{prop:gradient}
Let $\Pi({\boldsymbol \theta})$ be a $\nasp$ program and let $O$ be an observation such that $P_{\Pi({\boldsymbol \theta})}(O)>0$.
Let $p$ denote the probability of an atom $c=v$ in $\sigma^{nn}$, i.e., $p$ denotes \hbox{$P_{\Pi({\boldsymbol \theta})}(c=v)$}.  We have that\footnote{
$\frac{P_{\Pi({\boldsymbol \theta})}(I)}{P_{\Pi({\boldsymbol \theta})}(c=v)}$ and 
$\frac{P_{\Pi({\boldsymbol \theta})}(I)}{P_{\Pi({\boldsymbol \theta})}(c=v')}$
are still well-defined since the denominators have common factors in $P_{\Pi({\boldsymbol \theta})}(I)$.
}
\[
\frac{\partial log(P_{\Pi({\boldsymbol \theta})}(O))}{\partial p}
=
\frac{\sum\limits_{I :\ I\models O\atop I\models c=v} 
    \frac{P_{\Pi({\boldsymbol \theta})}(I)}{P_{\Pi({\boldsymbol \theta})}(c=v)}  
   - \sum\limits_{I, v' :\ I\models O\atop I\models c=v', v\neq v'} \frac{P_{\Pi({\boldsymbol \theta})}(I)}{P_{\Pi({\boldsymbol \theta})}(c=v')} }
{\sum\limits_{I :\ I\models O} P_{\Pi({\boldsymbol \theta})}(I)} .
\]
\end{prop}

Intuitively, the proposition tells us that each interpretation $I$ that satisfies $O$ tends to increase the value of $p$ if $I\models c=v$, and decrease the value of $p$ if $I\models c=v'$ such that $v'\neq v$. $\nasp$ internally calls {\sc clingo} to find all stable models $I$ of $\Pi({\boldsymbol \theta})$ that satisfy $O$ 
and uses PyTorch to obtain the probability of each atom $c=v$ in $\sigma^{nn}$.

\subsection{Experiment 1: Learning Digit Classification from Addition}

All experiments in Section~\ref{sec:learning} were done on 
Ubuntu 18.04.2 LTS with two 10-cores CPU Intel(R) Xeon(R) CPU E5-2640 v4 @ 2.40GHz and four GP104 [GeForce GTX 1080]. 

The digit addition problem is a simple example used in~\cite{manhaeve18deepproblog} to illustrate DeepProbLog's ability for both logical reasoning and deep learning. The task is, given a pair of digit images (MNIST) and their sum as the label, to let a neural network learn the digit classification of the input images.

The problem can be represented by $\nasp$ program $\Pi_{digit}$ in Example~\ref{ex:addition}. For comparison, we use the same dataset and the same structure of the neural network model used in \cite{manhaeve18deepproblog} to train the digit classifier $M_{digit}$ in $\Pi_{digit}$. For each pair of images denoted by $d_1$ and $d_2$ and their sum $n$, we construct the ASP constraint
$
~\leftarrow \no\ addition(d_1, d_2, n)~
$
as the observation $O$. The training target is to maximize $log(P_{\Pi_{digit}}(O))$. 

\begin{figure}[t!]
\centering
\caption{$\nasp$ Gradient Propagation}
\includegraphics[width=8.5cm]{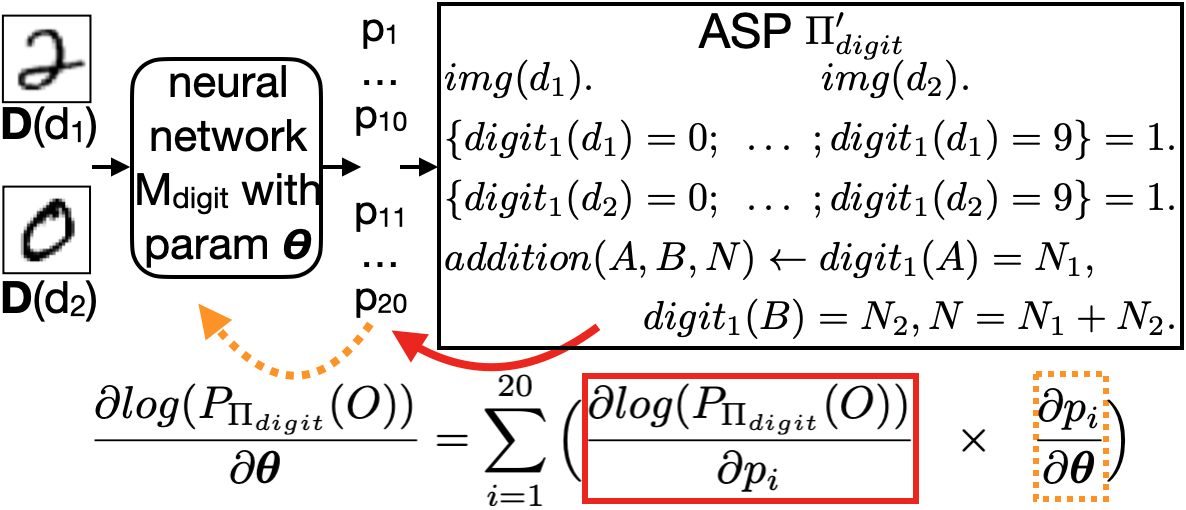}
\vspace{-0.4cm}
\label{fig:learning}
\end{figure}

Figure~\ref{fig:learning} shows how the forward and the backward propagations are done for 
$\nasp$ program $\Pi_{digit}$ in Example~\ref{ex:addition}.
The left-to-right direction is the forward computation of the neural network extended with the rule layer, whose output is the probability of the observation $O$. The right-to-left direction shows how the gradient from the rule layer is backpropagated further into the neural network by the chain rule to update all neural network parameters so as to find the parameter values that maximize the probability of the given observation.

\begin{figure}[ht!]
\begin{center}
\caption{{\small $\nasp$ v.s. DeepProbLog}}
\includegraphics[scale=0.3]{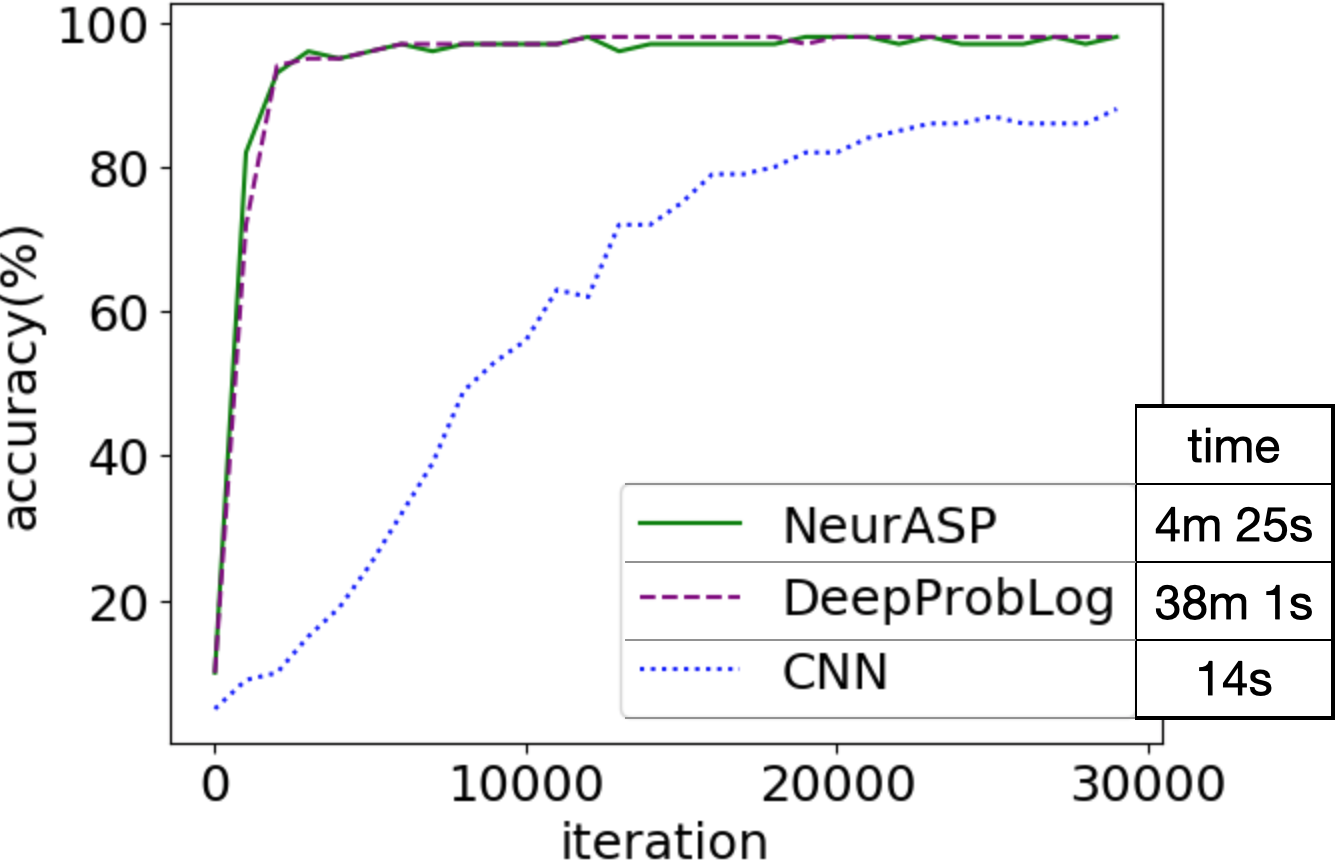}
\vspace{-0.4cm}
\label{fig:DeepLPMLNvsDeepProbLog}
\end{center}
\end{figure}

{
Figure~\ref{fig:DeepLPMLNvsDeepProbLog} shows the accuracy on the test data after each training iteration. The method CNN denotes the baseline used in \cite{manhaeve18deepproblog} where a convolutional neural network (with more parameters) is trained to classify the concatenation of the two images into the 19 possible sums. 
As we can see, the neural networks trained by $\nasp$ and DeepProbLog converge much faster than CNN and have almost the same accuracy at each iteration. 
However, $\nasp$ spends much less time on training compared to DeepProbLog. The time reported is for one epoch (30,000 iterations in gradient descent).
This is because DeepProbLog constructs an SDD (Sequential Decision Diagram) at each iteration for each training instance (i.e., each pair of images). This example illustrates that generating many SDDs could be more time-consuming than enumerating stable models in $\nasp$ computation. In general, there is a trade-off between the two methods and other examples may show the opposite behavior.
}


\NB{JL@Zhun: experiments comparing with DeepProbLog; note the time to construct SDD; show the graph; give analysis}

\subsection{Experiment 2: Learning How to Solve Sudoku}

In section~\ref{subsec:sudoku_inference}, we used a neural network $M_\mi{identify}$ to identify the numbers on a Sudoku board and used ASP rules to solve the Sudoku problem. 
In this section, we use a neural network to learn to solve Sudoku problems. The task is, given the {\em textual representation} of an unsolved Sudoku board (in the form of a $9\times 9$ matrix where an empty cell is represented by 0), to let a neural network learn to predict the solution of the Sudoku board. 

We use the neural network $M_{sol}$ from \cite{park18can} as the baseline. 
$M_{sol}$ is composed of 9 convolutional layers and a 1x1 convolution layer followed by softmax. \citeauthor{park18can} trained $M_{sol}$ using 1 million examples and achieved 70\% accuracy using an ``inference trick'': instead of predicting digits for all empty cells at once, which leads to a poor accuracy, the most probable grid-cell value was predicted one by one.

Since the current $\nasp$ implementation is not as scalable as neural network training, training on 1 million examples takes too long.  
Thus, we construct a dataset of 63,000 + 1000 $\langle config, label\rangle$ pairs for training and testing.
Using Park's method on this relatively small dataset, we observe that $M_{sol}$'s highest whole-board accuracy $Acc_\mi{sol}$\footnote{
The percentage of Sudoku examples that are correctly solved.} is only 29.1\% and $M_{sol}$'s highest grid-cell accuracy\footnote{
The percentage of grid cells having correct digits regardless whether the Sudoku solution is correct. 
} is only 89.3\% after 63 epochs of training.  

We get a better result by training $M_{sol}$ with the 
$\nasp$ program $\Pi_{sol}$. The program is almost the same as $\Pi_\mi{identify}$ in Section~\ref{subsec:sudoku_inference} except that it uses $M_\mi{sol}$ in place of $M_\mi{identify}$ and 
the first three rules of $\Pi_\mi{identify}$ are replaced with 
{\cblu 
\begin{lstlisting}
nn(sol(81, img), [1,2,3,4,5,6,7,8,9]).
a(R,C,N) :- sol(Pos,img,N), R=Pos/9, C=Pos\9.
\end{lstlisting}
}
because we do not have to assign the value {\tt empty} in  solving Sudoku.  

We trained $M_{sol}$ using $\nasp$ where the training target is to maximize the probability of all stable models that satisfy the observation. On the same test data, after 63 epochs of training, the highest whole-board accuracy of $M_{sol}$ trained this way is 66.5\% and the highest grid-cell accuracy is 96.9\% (In other words, we use rules only during training and not during testing). This indicates that including such structured knowledge sometimes helps the training of the neural network significantly. 

\BOCC
\begin{table}[ht]
\centering
\caption{{\cblu For solving sudoku, the grid-cell and the whole-board accuracies $Acc_{sol}$ of $M_{sol}$ under standard neural network training versus NeurASP program $\Pi_\mi{sol}$ on test data.}}
{\footnotesize 
\begin{tabular} {c| c| c| c| c}
\hline
\hline
Num of  & Grid-cell  & $Acc_{sol}$ of & Grid-cell &  $Acc_{sol}$ of  \\
Epochs  & Accuracy of      & Baseline &  Accuracy of  &  NeurASP w/  \\
      & Baseline &  &NeurASP w/  & $\Pi_\mi{sol}$   \\
      &  &  &  $\Pi_\mi{sol}$ &    \\
\hline
10  & 89.9\%    & 28.3\%  & 90.9\% & 25.9\%\\
20  & 87.76\%    & 21.5\%    & 94.1\% & 45.7\%\\
30  & 87.7\%   & 22.9\%    & 96.5\% & 62.9\%\\
40  & 88.2\% & 23.0\%    & 96.6\% & 64.4\%\\
50  & 88.2\%   & 23.6\%    & 96.8\% & 65.8\%\\
60  & 88.5\%   & 25.9\% & 96.9\% & 66.1\%\\
63  & 88.6\% & 23.0\%    & 96.9\% & 66.5\%\\

\hline
\hline
\end{tabular}
}
\label{tb:sudoku_solving}
\end{table}
\EOCC

\subsection{Experiment 3: Learning Shortest Path (SP)}

The experiment is about, given a graph and two points, finding the shortest path between them.
We use the dataset from \cite{xu18asemantic}, which was used to demonstrate the effectiveness of semantic constraints for enhanced neural network learning. Each example is a 4 by 4 grid $G = (V, E)$, where $\lvert V \rvert = 16, \lvert E \rvert = 24$.  The source and the destination nodes are randomly picked up, as well as 8 edges are randomly removed to increase the difficulty. The dataset is divided into 60/20/20 train/validation/test examples. 

The following $\nasp$ program
\footnote{$sp(X,g,true)$ means edge $X$ is in the shortest path. $sp(X,Y)$ means there is a path between nodes $X$ and $Y$ in the shortest path.}
\begin{lstlisting}
nn(sp(24, g), [true, false]). 
sp(0,1) :- sp(1,g,true).    
...      
sp(X,Y) :- sp(Y,X).
\end{lstlisting}
together with the union of the following 4 constraints defines the shortest path. 

\begin{lstlisting}
% [nr] 1. No removed edges should be predicted
:- sp(X,g,true), removed(X).

% [p] 2. Prediction must form a simple path, i.e.,
% the degree of each node must be either 0 or 2
:- X=0..15, #count{Y: sp(X,Y)} = 1.
:- X=0..15, #count{Y: sp(X,Y)} >= 3.

% [r] 3. Every 2 nodes in the prediction must be 
% reachable
reachable(X,Y) :- sp(X,Y).
reachable(X,Y) :- reachable(X,Z), sp(Z,Y).
:- sp(X,A), sp(Y,B), not reachable(X,Y).

% [o] 4. Predicted path should contain least edges 
:~ sp(X,g,true). [1, X]
\end{lstlisting}

\NB{comment is too long}

\BOC
\bi
\ii [{\bf [nr]}] No removed edges can be predicted.
\ii [{\bf [p]}] Prediction must form simple path(s).
\ii [{\bf [r]}] Every 2 nodes in the prediction must be reachable.
\ii [{\bf [o]}] Predicted path should contain minimum amount of edges.
\ei
\EOC
In this experiment, we trained the same neural network model $M_{sp}$ as in \cite{xu18asemantic}, a 5-layer Multi-Layer Perceptron (MLP), but with 4 different settings: (i) MLP only; (ii) together with $\nasp$ with the simple-path constraint {\bf (p)} (which is the only constraint used in \cite{xu18asemantic}); 
\footnote{A path is {\em simple} if every node in the path other than the source and the destination has only 1 incoming edge and only 1 outgoing edge.}
(iii) together with $\nasp$ with simple-path, reachability, and optimization constraints {\bf (p-r-o)}; and (iv) together with $\nasp$ with all 4 constraints {\bf (p-r-o-nr)}. 
\footnote{Other combinations are either meaningless (e.g., {\bf o}) or having similar results (e.g. {\bf p-r} is similar to {\bf p}).}

Table~\ref{tb:sp:Accuracy on Test Data} shows, after 500 epochs of training, the percentage of the predictions on the test data that satisfy each of the constraints {\bf p}, {\bf r}, and {\bf nr},  the path constraint (i.e., {\bf p-r}), the shortest path constraint (i.e., {\bf p-r-o-nr}), and the accuracy w.r.t. the ground truth. 

The accuracies for the first experiment (MLP Only) show that $M_{sp}$ was not trained well only by minimizing the cross-entropy loss of its prediction: 100-28.3 = 71.7\% of the predictions are not even a simple-path.

In the remaining experiments (MLP (x)), instead of minimizing the cross-entropy loss, our training target is changed to maximizing the probability of all stable models under certain constraints. 
The accuracies under the 2nd and 3rd experiments (MLP (p) and MLP (p-r-o) columns) are increased significantly, showing that 
(i) including such structured knowledge helps the training of the neural network and
(ii) the more structured knowledge included, the better $M_{sp}$ is trained under $\nasp$. 
Compared to the results from \cite{xu18asemantic}, $M_{sp}$ trained by $\nasp$ with the simple-path constraint {\bf p} (in the 2nd experiment MLP (p) column) obtains a similar accuracy on predicting the label (28.9\% v.s. 28.5\%) but a higher accuracy on predicting a simple-path (96.6\% v.s. 69.9\%).

In the 4th experiment (MLP (p-r-o-nr) column) where we added the constraint {\bf nr} saying that ``no removed edges can be predicted'',  the accuracies go  down. This is because the new constraint {\bf nr} is about randomly removed edges, changing from one example to another, which is hard to be generalized. 
\BOC
{\cred 
The results from the 2nd and the 3rd experiments show that the neural network is capable of predicting a path between 2 nodes. However, this accuracy drops to 30.1\% in the 4th experiment even though the ``quality'' of the stable models increases. One possible reason is that, comparing to the neural network trained with p-r-o-nr constraints, the neural network trained with p or p-r-o constraints uses more stable models to simulate a simpler task. Another reason is that the nr constraint introduces randomness from the removed edges, which may mislead the neural network from finding out a good feature to predict a path. 
}
\EOC

\vspace{-0.2cm}
\begin{table}[ht]
\caption{Shortest Path: Accuracy on Test Data: columns denote MLPs trained with different rules; each row represents the percentage of predictions that satisfy the constraints}
\centering
{\small 
\begin{tabular} {c| c| c| c| c}
\hline
\hline
Predictions & {MLP Only} & MLP & MLP & MLP\\
satisfying &  & (p) & (p-r-o) & (p-r-o-nr)\\
\hline
p & 28.3\% & 96.6\% & {\bf 100}\% & 30.1\%\\
r & 88.5\% & {\bf 100}\% &  {\bf 100}\% & 87.3\% \\
nr & 32.9\% & 36.3\% &  45.7\% & {\bf 70.5}\%\\
p-r &28.3\% & 96.6\% & {\bf 100}\% & 30.1\%\\
p-r-o-nr & 23.0\% & 33.2\% & {\bf 45.7}\% & 24.2\% \\
{\sl label (ground truth)} & 22.4\% & 28.9\% & {\bf 40.1}\% & 22.7\%\\
\hline
\hline
\end{tabular}
\label{tb:sp:Accuracy on Test Data}
}
\end{table}

\section{Related Work} \label{sec:related}

    
Recent years have observed the rising interests of combining perception and reasoning.
As mentioned, the work on DeepProbLog \cite{manhaeve18deepproblog} is closest to our work. Some differences are: (i) The computation of DeepProbLog relies on constructing circuits such as sequential decision diagrams (SDD) whereas we use an ASP solver internally. (ii) $\nasp$ employs expressive reasoning originating from answer set programming, such as defaults, aggregates, and optimization rules. This not only gives more expressive reasoning but also allows the more semantic-rich constructs as guide to learning. (iii) DeepProbLog requires each training data to be a single atom, while $\nasp$ allows each training data to be arbitrary propositional formulas.
\BOCC
{\cred (iv) DeepProbLog requires each training data to be a single atom, while $\nasp$ allows each training data to be arbitrary propositional formula so that more information can be used. 
\footnote{In practice, we turn arbitrary propositional formula into a set of ASP constraints so that the satisfiablity checking of the training data can be done in ASP.}
For instance, one can use a set of literals as a training data to represent all different observations at a time. 
}
\EOCC


Also related is using the semantic constraints to train neural networks better \cite{xu18asemantic}, but the constraints used in that work are simple propositional formulas whereas we use answer set programming language, in which it is more convenient to encode complex KR constraints.
Logic Tensor Network \cite{donadello17logic} is also related in that it uses neural networks to provide fuzzy values to atoms.

\BOC
Several other proposals were made to integrate statistical models and symbolic knowledge through loss functions, which determine the directions towards which the parameters of the statistical model are updated. \cite{hu16harnessing} proposed a general framework of using logic rules to enhance neural networks, where the neural network is seen as a ``student,'' while a ``teacher network'' uses a set of weighted first-order rules over the input-target space to regulate the prediction from the student network. The loss function does not only consider the distance between the prediction from the student network and the ground truth, but also the distance between predictions from the student network and from the teacher network. In this way, the student network is trained not only to fit the training data but also simulate the behavior of the teacher network, which is operated by a set of logic rules. Similarly, \cite{xu18asemantic} has investigated a semantic loss function that bridges between the vector output from neural networks and how much a given set of logical constraints are satisfied by the vector output. 
In such loose integration of logic rules and neural network through loss functions, the logical component is often seen as a blackbox --- the overall training process of the neural network is agnostic to the exact syntax and semantics of the logical component. 
\EOC 

Another approach 
is to embed logic rules in neural networks by representing logical connectives by mathematical operations and allowing the value of an atom to be a real number. For example, 
Neural Theorem Prover (NTP) \cite{rocktaschel17end} adopts the idea of dynamic neural module networks \cite{andreas16learning} to embed logic conjunction and disjunction in and/or-module networks. A proof-tree like end-to-end differentiable neural network is then constructed using Prolog's backward chaining algorithm with these modules. 
Another method that also constructs a proof-tree like neural network is TensorLog \cite{kathryn18tensorlog}, which uses matrix multiplication to simulate belief propagation that is tractable under the restriction that each rule is negation-free and can be transformed into a polytree. 

Graph neural network (GNN) \cite{kipf17semi} is a neural network model that is gaining more attention recently. Since a graph can encode objects and relations between objects, by learning message functions between the nodes, one can perform certain relational reasoning over the objects. For example, in~\cite{palm18recurrent}, it is shown that GNN can do well on Sudoku, but the input there is not an image but a textual representation. However, this is still restrictive compared to the more complex reasoning that KR formalisms provide.

Neuro-Symbolic Concept Learner \cite{mao19neuro} separates between visual perception and symbolic reasoning. It shows the data-efficiency by using only 10\% of the training data and achieving the state-of-the-art 98\% accuracy on CLEVR dataset.  
Our results are similar in the sense that using symbolic reasoning, we could use fewer data to achieve a high accuracy. 

$\nasp$ is similar to $\lpmln$ \cite{lee16weighted} in the sense that they are both probabilistic extensions of ASP and their semantics are defined by translations into ASP \cite{lee17lpmln}. $\lpmln$ allows any rules to be weighted, whereas $\nasp$ uses standard ASP rules.

\section{Conclusion} \label{sec:conclusion}
We showed that $\nasp$ can improve the neural network's perception result by applying reasoning over perceived objects and also can help neural network learn better by compensating the small size data with knowledge and constraints.
\NB{add probabilistic rules is not that difficult }
Since $\nasp$ is a simple integration of ASP with neural networks, it retains each of ASP and neural networks in individual forms, and can directly utilize the advances in each of them.  

The current implementation is a prototype and not highly scalable due to a naive computation of enumerating stable models. The future work includes how to make learning faster, and also 
analyzing the effects of the semantic constraints more systematically.

\section*{Acknowledgments}  
We are grateful to the anonymous referees for their useful comments. This work was partially supported by the National Science Foundation under Grant IIS-1815337.

\bibliographystyle{named}


\appendix
\onecolumn
\newpage

\section{Extend $\nasp$ With Probabilistic Rules}
{\cblu Multi-valued probabilistic programs are a fragment of $\lpmln$ programs that distinguishes between probabilistic rules and regular rules. 
We first present the definition of Multi-Valued Probabilistic Programs (MVPP) from~\cite{lee16weighted} with some modifications. Then we present the extended $\nasp$ with probabilistic rules, whose semantics is defined by a translation to MVPP. We assume that the reader is familiar with ASP-Core2 \cite{calimeri20asp}.}

\subsection{Multi-Valued Probabilistic Programs}
We assume a propositional signature $\sigma$ that is constructed from ``constants'' and their ``values.'' 
A {\em constant} $c$ is associated with a finite set $\i{Dom}(c)$, called the {\em domain} of $c$. 
The signature $\sigma$ is constructed from a finite set of constants, consisting of atoms $c\!=\!v$~
for every constant $c$ and every element $v$ in $\i{Dom}(c)$.
If the domain of~$c$ is $\{\false,\true\}$ then we say that~$c$ is {\em Boolean}, and abbreviate $c\mvis\true$ as $c$ and $c\mvis\false$ as~$\sneg c$.\footnote{The use of symbol $\sneg $\ \  is intentional; the semantics of $c\mvis \false$ works the same as strong negation.}
We assume that constants are divided into {\em probabilistic} constants and {\em non-probabilistic} constants. 
{\cblu 
By $\sigma_p$, we denote the set of atoms in $\sigma$ that are constructed from the probabilistic constants. }

\noindent{\bf Syntax: }
A {\em probabilistic rule} is of the form
\beq
p_1\!:\! c\mvis v_1 \ \ |\ \  p_2\!:\! c\mvis v_2 \ \ |\ \  \dots \ \ |\ \  p_n\!:\! c\mvis v_n
\eeq{n:eq:ad}
where $p_i$ are real numbers in $[0,1]$ (denoting probabilities) such that $\sum\limits_{i\in \{1,\dots,n \}} p_i=1$,
and $c$ is a probabilistic constant in $\sigma$, and
$\{v_1,\dots,v_n \}=\i{Dom}(c)$. If $\i{Dom}(c)=\{\false,\true\}$, rule $~~p_1: c \ \ | \ \ p_2: \sneg c~~$ can be abbreviated as
$~~
p_1: c
~~.$

A {\em Multi-Valued Probabilistic Program} is the union of $\Pi^{pr}$ and $\Pi^{asp}$, where 
$\Pi^{pr}$ consists of probabilistic rules~\eqref{n:eq:ad}, one for each probabilistic constant $c$ in $\sigma_p$, and 
$\Pi^{asp}$ consists of rules of the form
$~
\i{Head} \leftarrow \i{Body},
~$
following the rule format of ASP-Core2
where $\i{Head}$ contains no probabilistic constants.

\begin{example} \label{n:ex:coin}
Consider the game of flipping a coin where we win if we got $head$. Suppose the coin is biased and the probability of getting $head$ is 0.1, then this problem can be represented by the following MVPP program $\Pi_{coin}$ where $head$ is a probabilistic constant and $win$ is a non-probabilistic constant.
\begin{lstlisting}
0.1: head.
win :- head.
~win :- not win.
\end{lstlisting}
\end{example}


\medskip
\noindent{\bf Semantics: } Given an MVPP program $\Pi$, we obtain an ASP program $\Pi'$ from $\Pi$ by replacing each rule~\eqref{n:eq:ad} with
\[
1\{c\mvis v_1; \dots; c\mvis v_n\}1
\]
which means to choose only one atom from the set $\{c\mvis v_1,\dots, c\mvis v_n\}$.
In addition, $\Pi'$ contains the rule 
\[
\leftarrow  2\{c=v_1; \dots; c=v_n\}.
\]
for each non-probabilistic constant $c$ with $\i{Dom}(c) = \{v_1,\dots,v_n \}$. That is, non-probabilistic constants are allowed to have no values.

The stable models of $\Pi$ are defined as the stable models of $\Pi'$. 

{\cblu 
To define the probability of a stable model, we first define the probability of an atom $c\mvis v$ in $\sigma_p$. We know there must be exactly one probabilistic rule~\eqref{n:eq:ad} for each probabilistic constant $c$. Thus we can always find such a rule~\eqref{n:eq:ad} for any atom $c\mvis v$ in $\sigma_p$, and the probability of $c \mvis v_i$, denoted by $P_\Pi(c\mvis v_i)$, is defined as $p_i$ in rule~\eqref{n:eq:ad}.

The probability of a stable model $I$ of $\Pi$, denoted by $P_{\Pi}(I)$, is defined as the product of the probability of each atom $c=v$ in $I \cap \sigma_p$, divided by the number of stable models satisfied by $I \cap \sigma_p$. In the following equation, we use $I|_{\sigma_p}$ to denote $I \cap \sigma_p$, which is indeed the projection of $I$ onto $\sigma_p$. We also use $\i{Num}(I|_{\sigma_p},\Pi)$ to denote the number of stable models of $\Pi$ that satisfy $I|_{\sigma_p}$.
\[
P_{\Pi}(I) = \begin{cases}
\frac{\prod\limits_{c=v \in I|_{\sigma_p}} P_{\Pi}(c=v)}{\i{Num}(I|_{\sigma_p}, \Pi)}
 & \text{if $I$ is a stable model of $\Pi$;}\\
0 & \text{otherwise.}
\end{cases}
\]
An {\em observation} is a set of ASP constraints (i.e., rules of the form $~\bot \leftarrow \i{Body}$).
The probability of an observation $O$ is defined as
\[
  P_{\Pi}(O) = \sum\limits_{I\models  O}P_{\Pi}(I).
\]
The probability of the set $ {\bf O}=\{O_1,\dots, O_o \}$ of independent observations, where each $O_i$ is a set of ASP constraints, is defined as the product of the probability of each $O_i$:
\[
{\small 
P_{\Pi}({\bf O}) = \prod_{O_i\in {\bf O}}P_{\Pi}(O_i).
}
\]

\noindent{\bf Example~\ref{n:ex:coin} Continued: } The following ASP program is $\Pi_{coin}'$ (the ASP counter-part of $\Pi_{coin}$).
\begin{lstlisting}
win=true :- head=true.
win=false :- not win=true.

1{head=true; head=false}1.
:- 2{win=true; win=false}.
\end{lstlisting}
It has 2 stable models: $I_1 = \{head=\true, win=\true\}$ and $I_2 = \{head=\false, win=\false \}$, which are the stable models of $\Pi_{coin}$. There are 2 atoms in $\sigma_p$ and their probabilities are $P_{\Pi_{coin}}(head=\true)=0.1$ and $P_{\Pi_{coin}}(head=\false)=0.9$.
Then the probabilities of $I_1$ and $I_2$ can be computed as follows.
\[
P_{\Pi_{coin}}(I_1) = \frac{0.1}{1} = 0.1 ~~~~~~~~P_{\Pi_{coin}}(I_2) = \frac{0.9}{1} = 0.9
\]
And the probability of $O=\{\no\ win=\true \}$ is $P_{\Pi_{coin}}(O)) = P_{\Pi_{coin}}(I_2)=0.9$.
}

\subsection{Define $\nasp$ On a Translation to MVPP}
\noindent{\bf Syntax: }  We first define the notion of a neural atom to describe a neural network in a logic program. {\cblu Intuitively, a neural atom can be seen as the shorthand for a sequence of probabilistic rules whose atoms are defined by a syntactical translation from the neural atom and whose probabilities are the outputs of neural networks. }

We assume that neural network $M$ allows an arbitrary tensor as input whereas the output is a matrix in $\mathbb{R}^{e\times n}$, 
where $e$ is the number of random events predicted by the neural network, and $n$ is the number of possible outcomes for each random event. Each row of the matrix represents the probability distribution of the outcomes of each event. 
Given an input tensor ${\bf t}$, by $M({\bf t})$, we denote the output matrix of $M$.
$M({\bf t})[i,j]$ ($i\in\{1,\dots, e\}$, $j\in\{1,\dots,n\}$) is the probability at the $i$-th row and $j$-th column of the matrix. 
For example, in a neural network for MNIST digit classification, the input is a tensor representation of a digit image, $e$ is $1$, and $n$ is $10$. For a neural network that outputs a Boolean value for each edge in a graph, $e$ is the number of edges and $n$ is $2$. 


In $\nasp$, the neural network $M$ above can be represented by a {\em neural atom} of the form 
\beq
   nn(m(e, t), \left[v_1, \dots, v_n \right]), 
\eeq{n:eq:nn-atom}
where
(i) $nn$ is a reserved keyword to denote a neural atom; 
(ii) $m$ is an identifier (symbolic name) of the neural network $M$;
(iii) $t$ is a list of terms that serves as a ``pointer'' to an input data;  the mapping ${\bf D}$ is implemented by the external Python code that accepts $\nasp$ program as input, and can map $t$ to different data instances by iteratively loading from the dataset; this is useful for training;
%
(iv) $v_1, \dots, v_n$ represent all $n$ possible outcomes of each of the $e$ random events.
%
%
%

%
%
%

Each neural atom \eqref{n:eq:nn-atom} introduces propositional atoms of the form $c\mvis v$, where $c\in\{m_1(t), \dots, m_e(t)\}$ and $v\in\{v_1,\dots,v_n\}$. The output of the neural network provides the probabilities of the introduced atoms (defined in Section~\ref{ssec:semantics}).





\begin{example} \label{n:ex:digit}
Let $M_{digit}$ be a neural network that classifies an MNIST digit image. The input of $M_{digit}$ is (a tensor representation of) an image and the output is a matrix in $\mathbb{R}^{1\times 10}$. 
The neural network can be represented as the neural atom
\[
   nn( digit(1,d),\  [0,1,2,3,4,5,6,7,8,9])
\]
which introduces propositional atoms $digit_1(d)\mvis 0$, $digit_1(d)\mvis 1$, $\dots$, $digit_1(d)\mvis 9$.

Let $M_{sp}$ be another neural network for finding the shortest path in a graph with 24 edges. The input is a tensor encoding the graph and the start/end nodes of the path, and the output is a matrix in $\mathbb{R}^{24\times 2}$.
This neural network can be represented as the neural atom
\[
    nn( sp(24,g),\  [\true, \false]) .
\]
\end{example}

{\cblu 
A {\em $\nasp$ program} $\Pi$ is the union of $\Pi^{mvpp}$ and $\Pi^{nn}$ where $\Pi^{mvpp}$ is an MVPP program, and $\Pi^{nn}$ is a set of neural atoms.  Let $\sigma^{nn}$ be the set of all atoms $c\mvis v$ that is obtained from the neural atoms in $\Pi^{nn}$  as described above. We require that no atoms in $\sigma^{nn}$ appear in the probabilistic rules in $\Pi^{mvpp}$ or in $\i{Head}$ of each ASP rule $\i{Head}\ar\i{Body}$ in $\Pi^{mvpp}$.
}

We could allow schematic variables into $\Pi$, which are understood in terms of grounding as in standard answer set programs.  We find it convenient to use rules of the form 
\beq
   nn(m(e, t), \left[v_1, \dots, v_n \right]) \ar \i{Body}
\eeq{n:eq:nn-rule}
where $\i{Body}$ is either identified by $\top$ or $\bot$ during grounding so that \eqref{n:eq:nn-rule} can be viewed as an abbreviation of multiple (variable-free) neural atoms~\eqref{n:eq:nn-atom}.


\begin{example}\label{n:ex:addition}
An example $\nasp$ program $\Pi_{digit}$ is as follows, where  $d_1$ and $d_2$ are terms representing two images. Each image is classified by neural network $M_{digit}$ as one of the values in $\{0,\dots,9 \}$. The addition of two digit-images is the sum of their values. 
\[
{\small
\ba {l}
img(d_1). \\
img(d_2).\\
nn(digit(1, X), [0,1,2,3,4,5,6,7,8,9]) \leftarrow img(X).\\
addition(A,B,N) \leftarrow
    digit_1(A) \mvis N_1, 
    digit_1(B) \mvis N_2, \\
\hspace{3.05cm} N=N_1+N_2.
\ea
}
\]
The neural network $M_{digit}$ generates 10 probabilities for each image. The addition is applied once the digits are recognized and its probability is induced from the perception as we explain in the next section. 
\end{example}

{\cblu 
\noindent{\bf Semantics: } For any $\nasp$ program $\Pi=\Pi^{mvpp} \cup \Pi^{nn}$, we obtain its MVPP counterpart $\Pi'$ by replacing each neural atom $~nn(m(e, t), \left[v_1, \dots, v_n \right])~$ in $\Pi^{nn}$ with the set of probabilistic rules
\[
\ba {l}
p_{1,1}: m_1(t) \mvis v_1 ~|~ \dots ~|~  p_{1,n}: m_1(t) \mvis v_n\\
\dots \\
p_{e,1}: m_e(t) \mvis v_1 ~|~ \dots ~|~  p_{e,n}: m_e(t) \mvis v_n
\ea
\]
where $p_{i,j}$ denotes the probability of atom $m_i(t) \mvis v_j$. Recall that there is an external mapping ${\bf D}$ that turns $t$ into a specific input tensor of $M$, the value of $p_{i,j}$ is the neural network output $M({\bf D}(t))[i,j]$. 

The stable models of a $\nasp$ program $\Pi$ are defined as the stable models of its MVPP counterpart $\Pi'$. The probability of each stable model $I$ under $\Pi$ is defined as its probability under $\Pi'$.

\medskip
\noindent{\bf Example~\ref{n:ex:addition} Continued: } 
The following MVPP program is $\Pi'_{digit}$, i.e., the MVPP counterpart of $\Pi_{digit}$.
\[
{\small
\ba {l}
img(d_1). \\
img(d_2).\\
addition(A,B,N) \leftarrow
    digit_1(A) \mvis N_1, 
    digit_1(B) \mvis N_2, \\
\hspace{3.05cm} N=N_1+N_2. \\
p^{d_1}_{1,1}: digit_1(d_1)\mvis 0 ~|~ \dots ~|~ p^{d_1}_{1,10}: digit_1(d_1)\mvis 9 \\
p^{d_2}_{1,1}: digit_1(d_2)\mvis 0 ~|~ \dots ~|~ p^{d_2}_{1,10}: digit_1(d_2)\mvis 9
\ea
}
\]
Recall that $M_{digit}({\bf D}(d)) \in \mathbb{R}^{1\times 10}$ is the output matrix of $M_{digit}$ in Example~\ref{n:ex:addition} with input ${\bf D}(d)$. 
For $d\in \{ d_1, d_2\}$ and $j\in \{ 0,\dots,9\}$, the value of $p^{d}_{1,j}$ is the neural network output $M_{digit}({\bf D}(d))[1,j]$.
}

\BOC
\medskip
\noindent{\bf Example~\ref{n:ex:coin} Continued: } According to the semantics, $\Pi_{coin}'$ (the ASP counter-part of $\Pi_{coin}$) is as follows.
\begin{lstlisting}
1{head=t; head=f}1.
win=t :- head=t.
:- 2{win=t; win=f}.
\end{lstlisting}
There are 2 stable models of $\Pi_{coin}'$: $I_1 = \{head=\true, win=\true\}$ and $I_2 = \{head=\false\}$, which are thus the stable models of $\Pi_{coin}$. 
There are 2 total choices of $\Pi_{coin}$: $\{head=\true \}$ and $\{head=\false \}$, which are the stable models of $\Pi_{coin}^{ch}$ below.
\begin{lstlisting}
1{head=t; head=f}1.
\end{lstlisting}
Since, for each total choice $C$ of $\Pi_{coin}$, $\Pi_{coin}$ has exactly 1 stable model that satisfies $C$, $\Pi_{coin}$ is a 1-coherent MVPP program. The probabilities of $I_1$ and $I_2$ can be computed as follows.
\[
P_{\Pi_{coin}}(I_1) = \frac{1}{1} \times 0.1 = 0.1 ~~~~~~~~P_{\Pi_{coin}}(I_2) = \frac{1}{1} \times 0.9 = 0.9
\]
And the probability of $O=\{\no\ win=\true \}$ is $P_{\Pi_{coin}}(O)) = P_{\Pi_{coin}}(I_2)=0.9$.
\EOC

\newpage
\section{Proof of Proposition~\ref{prop:gradient}}
{\bf [Recall Proposition~\ref{prop:gradient}]}

Suppose $p_i$ is the probability of atom $c=v$, i.e., $p_i$ denotes $P_{\Pi({\boldsymbol \theta})}(c=v)$. 
\[
\frac{\partial log(P_{\Pi({\boldsymbol \theta})}(O))}{\partial p_i}
= \frac{\sum\limits_{I\models O\atop I\models c=v} \frac{P_{\Pi({\boldsymbol \theta})}(I)}{P_{\Pi({\boldsymbol \theta})}(c=v)}  
- \sum\limits_{I\models O\atop I\models c=v', v\neq v'} \frac{P_{\Pi({\boldsymbol \theta})}(I)}{P_{\Pi({\boldsymbol \theta})}(c=v')} }
{\sum\limits_{I\models O} P_{\Pi({\boldsymbol \theta})}(I)}
\]

\noindent{\bf [proof]}

\[
\ba {rl}
\frac{\partial log(P_{\Pi({\boldsymbol \theta})}(O))}{\partial p_i}
= & \frac{\partial log(P_{\Pi({\boldsymbol \theta})}(O))}{\partial P_{\Pi({\boldsymbol \theta})}(c=v)} \\
\\
= & \frac{1}{P_{\Pi({\boldsymbol \theta})}(O)} \times \frac{\partial P_{\Pi({\boldsymbol \theta})}(O)}{\partial P_{\Pi({\boldsymbol \theta})}(c=v)} \\
\\
\multicolumn{2}{l}{\text{(since $P_{\Pi({\boldsymbol \theta})}(O) = \sum\limits_{I\models  O}P_{\Pi({\boldsymbol \theta})}(I)$)}}\\
= & \frac{1}
{\sum\limits_{I\models O} P_{\Pi({\boldsymbol \theta})}(I)} \times \frac{\partial \sum\limits_{I\models  O}P_{\Pi({\boldsymbol \theta})}(I)}{\partial P_{\Pi({\boldsymbol \theta})}(c=v)} \\
\\
\multicolumn{2}{l}{\text{(since (i) $P_{\Pi({\boldsymbol \theta})}(I) = 0$ if $I$ is not a stable model of $\Pi({\boldsymbol \theta})$ and (ii) any stable model $I$ of $\Pi({\boldsymbol \theta})$ must satisfy $c=v^*$ for some $v^*$)}}\\
= & \frac{1}
{\sum\limits_{I\models O} P_{\Pi({\boldsymbol \theta})}(I)} \times \Big( \frac{\partial \sum\limits_{I \text{ is a stable model of } \Pi({\boldsymbol \theta})\atop {I\models  O\atop I\vDash c=v}}P_{\Pi({\boldsymbol \theta})}(I)}{\partial P_{\Pi({\boldsymbol \theta})}(c=v)} + 
\frac{\partial \sum\limits_{I, v' \mid I \text{ is a stable model of } \Pi({\boldsymbol \theta})\atop {I\models  O\atop I\vDash c=v', v\neq v'}}P_{\Pi({\boldsymbol \theta})}(I)}{\partial P_{\Pi({\boldsymbol \theta})}(c=v)} \Big) \\
\\
\multicolumn{2}{l}{\text{(since for any stable model $I$ of $\Pi({\boldsymbol \theta})$, $P_{\Pi({\boldsymbol \theta})}(I) = \frac{\prod\limits_{c^*=v^* \in I|_{\sigma_m}} P_{\Pi({\boldsymbol \theta})}(c^*=v^*)}{Num(I|_{\sigma_m}, \Pi({\boldsymbol \theta}))}$)}}\\
= & \frac{1}
{\sum\limits_{I\models O} P_{\Pi({\boldsymbol \theta})}(I)} \times \Big( \frac{\partial \sum\limits_{I \text{ is a stable model of } \Pi({\boldsymbol \theta})\atop {I\models  O\atop I\vDash c=v}} 
\frac{\prod\limits_{c^*=v^* \in I|_{\sigma_m}} P_{\Pi({\boldsymbol \theta})}(c^*=v^*)}{Num(I|_{\sigma_m}, \Pi({\boldsymbol \theta}))}
}{\partial P_{\Pi({\boldsymbol \theta})}(c=v)} + \frac{\partial \sum\limits_{I \text{ is a stable model of } \Pi({\boldsymbol \theta})\atop {I\models  O\atop I\vDash c=v', v\neq v'}}
\frac{\prod\limits_{c^*=v^* \in I|_{\sigma_m}} P_{\Pi({\boldsymbol \theta})}(c^*=v^*)}{Num(I|_{\sigma_m}, \Pi({\boldsymbol \theta}))}
}{\partial P_{\Pi({\boldsymbol \theta})}(c=v)} \Big) \\
\\
= & \frac{1}
{\sum\limits_{I\models O} P_{\Pi({\boldsymbol \theta})}(I)} \times 
\Big( 
\sum\limits_{I \text{ is a stable model of } \Pi({\boldsymbol \theta})\atop {I\models  O\atop I\vDash c=v}} 
\frac{\partial  
\frac{\prod\limits_{c^*=v^* \in I|_{\sigma_m}} P_{\Pi({\boldsymbol \theta})}(c^*=v^*)}{Num(I|_{\sigma_m}, \Pi({\boldsymbol \theta}))}
}{\partial P_{\Pi({\boldsymbol \theta})}(c=v)} + 
\sum\limits_{I \text{ is a stable model of } \Pi({\boldsymbol \theta})\atop {I\models  O\atop I\vDash c=v', v\neq v'}}
\frac{\partial 
\frac{\prod\limits_{c^*=v^* \in I|_{\sigma_m}} P_{\Pi({\boldsymbol \theta})}(c^*=v^*)}{Num(I|_{\sigma_m}, \Pi({\boldsymbol \theta}))}
}{\partial P_{\Pi({\boldsymbol \theta})}(c=v)} \Big) \\
\\
= & \frac{1}
{\sum\limits_{I\models O} P_{\Pi({\boldsymbol \theta})}(I)} \times 
\Big( 
\sum\limits_{I \text{ is a stable model of } \Pi({\boldsymbol \theta})\atop {I\models  O\atop I\vDash c=v}} 
\frac{ 
\frac{\prod\limits_{c^*=v^* \in I|_{\sigma_m}} P_{\Pi({\boldsymbol \theta})}(c^*=v^*)}{Num(I|_{\sigma_m}, \Pi({\boldsymbol \theta}))}
}{P_{\Pi({\boldsymbol \theta})}(c=v)} + 
\sum\limits_{I \text{ is a stable model of } \Pi({\boldsymbol \theta})\atop {I\models  O\atop I\vDash c=v', v\neq v'}}
\frac{\partial 
\frac{\prod\limits_{c^*=v^* \in I|_{\sigma_m}} P_{\Pi({\boldsymbol \theta})}(c^*=v^*)}{Num(I|_{\sigma_m}, \Pi({\boldsymbol \theta}))}
}{\partial P_{\Pi({\boldsymbol \theta})}(c=v')} \times \frac{\partial P_{\Pi({\boldsymbol \theta})}(c=v')}{\partial P_{\Pi({\boldsymbol \theta})}(c=v)} \Big)\\
\\
\multicolumn{2}{l}{\text{(since $P_{\Pi({\boldsymbol \theta})}(c=v') = 1- P_{\Pi({\boldsymbol \theta})}(c=v) - \dots$)}}\\
= & \frac{1}
{\sum\limits_{I\models O} P_{\Pi({\boldsymbol \theta})}(I)} \times 
\Big( 
\sum\limits_{I \text{ is a stable model of } \Pi({\boldsymbol \theta})\atop {I\models  O\atop I\vDash c=v}} 
\frac{ 
\frac{\prod\limits_{c^*=v^* \in I|_{\sigma_m}} P_{\Pi({\boldsymbol \theta})}(c^*=v^*)}{Num(I|_{\sigma_m}, \Pi({\boldsymbol \theta}))}
}{P_{\Pi({\boldsymbol \theta})}(c=v)} + 
\sum\limits_{I \text{ is a stable model of } \Pi({\boldsymbol \theta})\atop {I\models  O\atop I\vDash c=v', v\neq v'}}
\frac{
\frac{\prod\limits_{c^*=v^* \in I|_{\sigma_m}} P_{\Pi({\boldsymbol \theta})}(c^*=v^*)}{Num(I|_{\sigma_m}, \Pi({\boldsymbol \theta}))}
}{P_{\Pi({\boldsymbol \theta})}(c=v')} \times -1 \Big) \\
\\
=& \frac{1 }
{\sum\limits_{I\models O} P_{\Pi({\boldsymbol \theta})}(I)} \times 
\Big(
\sum\limits_{I\models O\atop I\models c=v} \frac{P_{\Pi({\boldsymbol \theta})}(I)}{P_{\Pi({\boldsymbol \theta})}(c=v)}  
- \sum\limits_{I\models O\atop I\models c=v', v\neq v'} \frac{P_{\Pi({\boldsymbol \theta})}(I)}{P_{\Pi({\boldsymbol \theta})}(c=v')} \Big)
\ea
\]

\section{More on Sudoku Experiments}

[[ Adam ]]

\section{Detailed Description of Learning Algorithms for $\nasp$}
Consider a $\nasp$ program $\Pi({\boldsymbol \theta})$ and a set ${\bf O}$ of observations such that $P_{\Pi({\boldsymbol \theta})}(O)>0$ for each $O \in {\bf O}$. 
The task is to find $\hat{\boldsymbol \theta}$ that maximizes the log-likelihood of observations ${\bf O}$ under program $\Pi({\boldsymbol \theta})$, i.e.,  
\[
\hat{\boldsymbol \theta} \in \argmax_{{\boldsymbol \theta}}~ log( P_{\Pi({\boldsymbol \theta})}({\bf O})), 
\]
which is equivalent to 
\[
\hat{\boldsymbol \theta} \in \argmax_{{\boldsymbol \theta}}~ \sum\limits_{O \in {\bf O}} log( P_{\Pi({\boldsymbol \theta})}(O)).
\]
Let ${\bf p}$ denote the probabilities of the atoms in $\sigma^{nn}$. Since ${\bf p}$ is indeed the outputs of the neural networks in $\Pi({\boldsymbol \theta})$, we can compute the gradient of ${\bf p}$ w.r.t. ${\boldsymbol \theta}$ through back-propagation. 
Then the gradients of $\sum\limits_{O \in {\bf O}} log( P_{\Pi({\boldsymbol \theta})}(O))$ w.r.t. ${\boldsymbol \theta}$ is
\[
\ba {rl}
\frac{\partial \sum\limits_{O \in {\bf O}} log( P_{\Pi({\boldsymbol \theta})}(O))}{\partial {\boldsymbol \theta}} 
& = \sum\limits_{O \in {\bf O}} \frac{\partial log(P_{\Pi({\boldsymbol \theta})}(O))}{\partial {\bf p}} \times \frac{\partial {\bf p}}{\partial {\boldsymbol \theta}}
\ea
\]
where $\frac{\partial {\bf p}}{\partial {\boldsymbol \theta}}$ can be computed through the usual neural network back-propagation, while $\frac{\partial log(P_{\Pi({\boldsymbol \theta})}(O))}{\partial p}$ for each $p \in {\bf p}$ can be computed as follows.

\medskip
\noindent{\bf Proposition~\ref{prop:gradient}~~~}
Let $p$ denote the probability of atom $c=v$, i.e., $p$ denotes $P_{\Pi({\boldsymbol \theta})}(c=v)$. 
\[
\frac{\partial log(P_{\Pi({\boldsymbol \theta})}(O))}{\partial p}
= \frac{\sum\limits_{I\models O\atop I\models c=v} \frac{P_{\Pi({\boldsymbol \theta})}(I)}{P_{\Pi({\boldsymbol \theta})}(c=v)}  
- \sum\limits_{I\models O\atop I\models c=v', v\neq v'} \frac{P_{\Pi({\boldsymbol \theta})}(I)}{P_{\Pi({\boldsymbol \theta})}(c=v')} }
{\sum\limits_{I\models O} P_{\Pi({\boldsymbol \theta})}(I)}
\]

Algorithm~\ref{alg:learn:exact} shows how to update the value of ${\boldsymbol \theta}$ to maximize the log-likelihood of observations ${\bf O}$ under $\Pi({\boldsymbol \theta})$.

\begin{algorithm}[ht!]
{\footnotesize
\noindent {\bf Input: }
\begin{enumerate}
\item $\Pi({\boldsymbol \theta})$: a $\nasp$ program (under signature $\sigma$) with parameters ${\boldsymbol \theta}$
\item ${\bf O}$: a set of observations $\{O_1,\dots,O_n \}$ where each $O_i$ is a set of ASP constraints such that $P_{\Pi({\boldsymbol \theta})}(O)>0$
\item ${\bf D}$: a set of mappings $\{{\bf D_1},\dots, {\bf D_n} \}$ where each ${\bf D_i}$ is associated with $O_i$ and maps terms to input tensors of neural networks in $\Pi({\boldsymbol \theta})$
\item lr: a real number denoting learning rate
\item epoch: a positive integer denoting the number of epochs
\end{enumerate}

\noindent {\bf Output: } 
\begin{enumerate}
    \item ${\hat {\boldsymbol \theta}}$: the updated parameters such that
$
{\hat {\boldsymbol \theta}} \in \argmax_{\boldsymbol \theta}~ log(P_{\Pi({\boldsymbol \theta})}({\bf O}))
$
\end{enumerate}

\noindent {\bf Procedure:}
\begin{enumerate}
\item Repeat for epoch number of times:
\begin{enumerate}
    \item For $O_i$ in ${\bf O}$:
    \begin{enumerate}
        \item Compute ${\bf p}$, i.e., the outputs of the neural networks in $\Pi({\boldsymbol \theta})$ according to ${\bf D_i}$
        \item Compute $\frac{\partial {\bf p}}{\partial {\boldsymbol \theta}}$ by back-propagation
        \item Find all stable models ${\bf I}_{O_i}$ of $\Pi({\boldsymbol \theta})$ that satisfy $O_i$ (by calling {\sc clingo} on $\Pi' \cup O_i$)
        \item gradients = $gradientsSM$($\Pi({\boldsymbol \theta})$, ${\bf p}$, ${\bf I}_{O_i}$) (Here, gradients is indeed $\frac{\partial log(P_{\Pi({\boldsymbol \theta})}(O_i))}{\partial {\bf p}}$)
        \item ${\boldsymbol \theta}$ = ${\boldsymbol \theta}$ + lr * gradients * $\frac{\partial {\bf p}}{\partial {\boldsymbol \theta}}$
        
    \end{enumerate}
\end{enumerate}
\item return ${\boldsymbol \theta}$

\end{enumerate}
\caption{$\nasp$ weight learning by exact computation}
\label{alg:learn:exact}
}
\end{algorithm}

Algorithm~\ref{alg:learn:exact} uses the function
$gradientsSM$, which is defined in Algorithm~\ref{alg:gradientsSM} below.

\begin{algorithm}[ht!]
{\footnotesize
\noindent {\bf Input: }
\begin{enumerate}
\item $\Pi({\boldsymbol \theta})$: a $\nasp$ program (under signature $\sigma$) with parameters ${\boldsymbol \theta}$
\item ${\bf p}$: the probabilities of the atoms in $\sigma^{nn}$
\item ${\bf I}$: a set of stable models of $\Pi({\boldsymbol \theta})$, where the summation of their probabilities is to be maximized 
\end{enumerate}

\noindent {\bf Output: } 
\begin{enumerate}
    \item $\frac{\partial log(\sum\limits_{I_i\in {\bf I}}P_{\Pi({\boldsymbol \theta})}(I_i))}{\partial {\bf p}}$: the gradients of the log-likelihood of ${\bf I}$ w.r.t. ${\bf p}$
\end{enumerate}

\noindent {\bf Procedure:}
\begin{enumerate}
\item If $|{\bf I}|=1$:
\begin{enumerate}
    \item for each $p \in {\bf p}$, where $p$ denotes the probability of one atom $c=v\in \sigma^{nn}$:
    \begin{itemize}
        \item if $I$ contains $c=v$, gradient($p$) = $\frac{1}{p}$
        \item else, $I$ must contain $c=v'$ for some $v'\neq v$, and gradient($p$) = -$\frac{1}{P_{\Pi({\boldsymbol \theta})}(c=v')}$
    \end{itemize}
\end{enumerate}
\item If $|{\bf I}|\geq 2$:
\begin{enumerate}
    \item compute $P_{\Pi({\boldsymbol \theta})}(I)$ for each $I$ in ${\bf I}$
    \item denominator = $\sum\limits_{I\in {\bf I}} P_{\Pi({\boldsymbol \theta})}(I)$
    \item for each $p \in {\bf p}$, where $p$ denotes the probability of one atom $c=v\in \sigma^{nn}$:
    \begin{enumerate}
    \item numerator = 0
    \item for each $I\in {\bf I}$:
    \begin{itemize}
        \item if $I$ contains $c=v$, numerator += $\frac{1}{p}  P_{\Pi({\boldsymbol \theta})}(I)$
        \item else, $I$ must contain $c=v'$ for some $v'\neq v$, and numerator -= $\frac{1}{P_{\Pi({\boldsymbol \theta})}(c=v')}  P_{\Pi({\boldsymbol \theta})}(I)$
    \end{itemize}
    \item gradient($p$) $= \frac{\text{numerator}}{\text{denominator}}$
    \end{enumerate}
\end{enumerate}
\item return [gradient($p$) for $p$ in ${\bf p}$]
\end{enumerate}

\caption{$gradientsSM$: compute gradients of the probability of a set of stable models}
\label{alg:gradientsSM}
}
\end{algorithm}

Algorithm~\ref{alg:learn:sampling} is almost the same as Algorithm~\ref{alg:learn:exact} except that, in step 1-(a)-iii, instead of finding all stable models that satisfy $O_i$, it randomly samples num\_of\_samples stable models that satisfy $O_i$ according to their probability distribution. The function $sampleSM$ used in Algorithm~\ref{alg:learn:sampling} to sample stable models is defined in Algorithm~\ref{alg:learn:sample_obs}.

\begin{algorithm}[ht!]
{\footnotesize
\noindent {\bf Input: }
\begin{enumerate}
\item $\Pi({\boldsymbol \theta})$: a $\nasp$ program (under signature $\sigma$) with parameters ${\boldsymbol \theta}$
\item ${\bf O}$: a set of observations $\{O_1,\dots,O_n \}$ where each $O_i$ is a set of ASP constraints such that $P_{\Pi({\boldsymbol \theta})}(O)>0$
\item ${\bf D}$: a set of mappings $\{{\bf D_1},\dots, {\bf D_n} \}$ where each ${\bf D_i}$ is associated with $O_i$ and maps terms to input tensors of neural networks in $\Pi({\boldsymbol \theta})$
\item num\_of\_samples: the number of sample stable models generated for each $O_i$ in each iteration
\item lr: a real number denoting learning rate
\item epoch: a positive integer denoting the number of epochs
\end{enumerate}

\noindent {\bf Output: } 
\begin{enumerate}
    \item ${\hat {\boldsymbol \theta}}$: the updated parameters such that
$
{\hat {\boldsymbol \theta}} \in \argmax_{\boldsymbol \theta}~ log(P_{\Pi({\boldsymbol \theta})}({\bf O}))
$
\end{enumerate}

\noindent {\bf Procedure:}
\begin{enumerate}
\item Repeat for epoch number of times:
\begin{enumerate}
    \item For $O_i$ in ${\bf O}$:
    \begin{enumerate}
        \item Compute ${\bf p}$, i.e., the outputs of the neural networks in $\Pi({\boldsymbol \theta})$ according to ${\bf D_i}$
        \item Compute $\frac{\partial {\bf p}}{\partial {\boldsymbol \theta}}$ by back-propagation
        \item Sample num\_of\_samples stable models of $\Pi({\boldsymbol \theta})$ that satisfy $O_i$ according to their probability distribution:
        \[
        {\bf I} = sampleSM(\Pi({\boldsymbol \theta}), O_i, \text{num\_of\_samples})\]
        \item gradients = $gradientsSM$($\Pi({\boldsymbol \theta})$, ${\bf p}$, ${\bf I}$) (Here, gradients is indeed an approximate of $\frac{\partial log(P_{\Pi({\boldsymbol \theta})}(O_i))}{\partial {\bf p}}$)
        \item ${\boldsymbol \theta}$ = ${\boldsymbol \theta}$ + lr * gradients * $\frac{\partial {\bf p}}{\partial {\boldsymbol \theta}}$
        
    \end{enumerate}
\end{enumerate}
\item return ${\boldsymbol \theta}$

\end{enumerate}
\caption{$\nasp$ weight learning by sampling stable models}
\label{alg:learn:sampling}
}
\end{algorithm}

\begin{algorithm}[ht!]
{\footnotesize
\noindent {\bf Input: }
\begin{enumerate}
\item $\Pi({\boldsymbol \theta})$: a $\nasp$ program (under signature $\sigma$) with parameters ${\boldsymbol \theta}$
\item $O$: an observation in the form of a set of ASP constraints
\item num\_of\_samples: the number of sample stable models generated for $O$
\end{enumerate}

\noindent {\bf Output: } 
\begin{enumerate}
    \item ${\bf I}$: a list of stable models of $\Pi({\boldsymbol \theta}) \cup O$ such that $|{\bf I}| \geq$ num\_of\_samples and the probability distribution of ${\bf I}$ follows the distribution defined by $\Pi({\boldsymbol \theta})$
\end{enumerate}

\noindent {\bf Procedure:}
\begin{enumerate}
\item SM = []
\item obtain $\Pi'$ from $\Pi({\boldsymbol \theta})$ according to the semantics
\item while True: do
\begin{enumerate}
    \item obtain an ASP program $P$ from $\Pi'$ by randomly replacing each choice rule
    \[
    \{c=v_1 ~;~ \dots ~;~ c=v_n\} =1
    \]
    in $\Pi'$ with a fact ``$c=v_i.$'' according to the probability distribution $\langle P_{\Pi({\boldsymbol \theta})}(c=v_1), \dots, P_{\Pi({\boldsymbol \theta})}(c=v_n)\rangle$;
    \item generate the set $S$ of all stable models of $P\cup O$ (by calling {\sc clingo} on $P\cup O$);
    \item append each element in $S$ to SM;
    \item if $|SM|\geq $ num\_of\_samples: break the loop;
\end{enumerate}
\item return SM
\end{enumerate}
\caption{$sampleSM$: sample num\_of\_samples stable models that satisfy $O$}
\label{alg:learn:sample_obs}
}
\end{algorithm}
\end{document}